\documentclass[pdflatex,sn-mathphys-num, oneside]{sn-jnl}
\usepackage{graphicx}%
\usepackage{multirow}%
\usepackage{amsmath,amssymb,amsfonts}%
\usepackage{amsthm}%
\usepackage{mathrsfs}%
\usepackage[title]{appendix}%
\usepackage{xcolor}%
\usepackage{textcomp}%
\usepackage{manyfoot}%
\usepackage{booktabs}%
\usepackage{algorithm}%
\usepackage{algorithmicx}%
\usepackage{algpseudocode}%
\usepackage{listings}%
\usepackage{flexisym}
\usepackage{hyperref}
\usepackage[capitalise]{cleveref}
\usepackage{spverbatim}
\usepackage[most]{tcolorbox}
\usepackage{mdframed}
\usepackage{framed}
\usepackage{hhline}
\usepackage{xcolor}

\theoremstyle{thmstyleone}%
\theoremstyle{thmstyletwo}%

\theoremstyle{thmstylethree}%

\begin{document}

\title[Article Title]{Measuring (some aspects of) the metacognition of AI}






\author*[1,3]{\fnm{} \sur{Richard Servajean}}\email{rservajean@yahoo.fr}\equalcont{These authors contributed equally to this work.}
\author[2]{\fnm{} \sur{Philippe Servajean}}\email{philippe\_gi@hotmail.fr}\equalcont{These authors contributed equally to this work.}

\affil*[1]{\orgdiv{Center for Brain Science}, \orgname{RIKEN}, \orgaddress{\city{Saitama}, \country{Japan}}}
\affil[2]{\orgdiv{Department of Psychology}, \orgname{Paul-Valéry University}, \orgaddress{\city{Montpellier}, \country{France}}}
\affil[3]{Current address: \orgname{National Institute of Advanced Industrial Science and Technology (AIST)}, \orgaddress{\city{Tokyo}, \country{Japan}}}

\abstract{\unboldmath A robust decision-making process must take into account uncertainty, especially when the choice involves inherent risks. Because artificial intelligence (AI) systems are increasingly integrated into decision-making workflows, managing uncertainty relies more and more on the metacognitive capabilities of these systems; i.e, their ability to assess the reliability of and regulate their own decisions. Hence, it is crucial to employ robust methods to measure the metacognitive abilities of AI. This paper is primarily a methodological contribution that highlights a key limitation of commonly used measures of AI \textit{metacognitive sensitivity}—the ability to generate confidence ratings that distinguish correct from incorrect responses. We then draw attention to the meta-$d'$ framework, a well-established approach from psychology and neuroscience designed to address this limitation. Moreover, we propose to leverage \textit{signal detection theory} (SDT) to measure the ability of AIs to spontaneously regulate their decisions based on uncertainty and risk. To demonstrate the practical utility of these psychophysical frameworks, we conduct two series of experiments on three large language models (LLMs)—GPT-5, DeepSeek-V3.2-Exp, and Mistral-Medium-2508.}



\maketitle

\newpage

\section{Introduction}\label{sec1}

It is widely acknowledged that artificial intelligence (AI) is transforming our societies \cite{makridakis2017forthcoming, stone2022artificial}, in fields as diverse as education \cite{hwang2023review} and protein structure prediction \cite{jumper2021highly}. In particular, since the release of GPT-3.5 in 2022 \cite{openai2022}, the massive adoption of conversational chatbots based on large language models (LLMs) by the public in their daily lives \cite{explodingtopics2025chatbot} has reignited debates about the opportunities and risks posed by AI \cite{weidinger2021ethical, rillig2023risks}. This has motivated many researchers to test the current capabilities of AI systems across a broad range of tasks, and specifically those of conversational chatbots \cite{chang2024survey}. 

As AI systems are increasingly deployed across society, we identify three ways in which they can be incorporated into a decision process:
\begin{enumerate}
\item \textbf{AI decision-making:} The AI system operates alone without any human intervention. An example of this would be autonomous driving at Level 4 or 5 (as defined by the SAE J3016 standard \cite{on2021taxonomy}), denoting systems that require no human oversight.

\item \textbf{Human-AI joint decision-making:} the AI system and humans take a decision as co-agents. One example is a setting in which the final decision results from a combination of human and AI judgments \cite{steyvers2022bayesian}.


\item \textbf{AI-assisted decision-making:} the AI system makes recommendations that are ultimately approved, refined, or ignored by a human. An example of this would be AI-powered medical diagnostics, where the system flags potential abnormalities in scans and forwards them to a radiologist who is responsible for making the final medical conclusion \cite{castilla2025external}.

\end{enumerate}

It should be noted that these categories only serve as primary landmarks along a broader continuum, reflecting the vast diversity of possible human-AI interaction patterns. See Ref.~\cite{gomez2025human} for a review of human-AI interaction patterns identified in the literature. Regardless, an effective decision-making process requires taking \textit{uncertainty} into account \cite{platt2008risky, li2025beyond}. Here, uncertainty concerns the probability that a judgment or a prediction (e.g., regarding the outcome of a choice) is correct. When uncertainty and risks are high, the most effective strategy is often to suspend judgment and prioritize further data collection before making the final decision (see the \textit{exploration-exploitation dilemma} \cite{berger2014exploration, mehlhorn2015unpacking}). Furthermore, in some decision-making contexts, the available options may differ considerably in the magnitude of their associated risks and/or benefits. In such cases, accounting for uncertainty becomes critically important. For instance, even a marginal amount of uncertainty regarding a catastrophic outcome can be enough to warrant an alternative choice that is more uncertain but offers lower risks in the failure scenario. Finally, in some multi-agent (human-human, human-AI or AI-AI) joint decision-making, quantifying the uncertainty of each agent's prediction provides a principled way to arbitrate between divergent forecasts \cite{bahrami2010optimally, steyvers2022bayesian, nguyen2025joint, bhattacharyya2024towards, koriat2012two}. 
Overall, note that researchers from fields as diverse as game theory \cite{von2007theory}, computational neuroscience \cite{parr2022active} and machine learning \cite{vo2019deep} have investigated the question of finding optimal strategies within uncertain contexts.

Crucially, managing uncertainty poses unique challenges, depending on how AI systems are incorporated into decision processes. In the \textit{AI decision-making} scenario, AI systems operate autonomously. The challenges therefore concern the ability of AI systems to estimate uncertainty and adjust their decisions accordingly. Here, the role of humans is limited to providing preliminary information, such as acceptable risk thresholds. In contrast, \textit{human-AI joint} and \textit{AI-assisted} decision-making can involve an additional challenge: the effective communication of uncertainty information between humans and AI \cite{steyvers2025improving, steyvers2025large, li2025beyond, steyvers2025metacognition, lee2025metacognitive}. In these collaborative scenarios, AI systems must not only evaluate uncertainty, but also convey it transparently to humans (ideally accompanied by explanations about the sources of that uncertainty in order to determine whether it can be reduced, and, if so, how). For example, this could allow humans to strategically allocate verification efforts only to instances where the AI system identifies—and reports—its outputs as unreliable.


Against this backdrop, the current work examines these uncertainty-related challenges by focusing specifically on LLMs, whose integration into decision-making workflows has become increasingly prevalent. More precisely, we investigate two specific capabilities of LLMs. First, the ability of LLMs to assess and report the reliability of their own predictions or judgments. Second, the ability of LLMs to spontaneously adjust their decisions based on this estimated uncertainty and risk configuration. As previously mentioned, it is worth noting that the relevance of these capabilities varies significantly depending on the decision-making scenario at play. Specifically, the human-AI joint and AI-assisted decision-making scenarios involve both the ability to estimate and report uncertainty. In contrast, the AI decision-making scenario, where the AI operates autonomously, would require only assessing uncertainty without reporting it. Finally, the ability to adjust decisions based on uncertainty and risk would be mainly involved in the AI decision-making scenario.


Two core dimensions underlie the ability of LLMs to assess and report the reliability of their own outcomes: \textit{metacognitive sensitivity}  and \textit{metacognitive calibration} (also known as type 2 sensitivity and type 2 bias, respectively). The concepts of metacognitive sensitivity and metacognitive calibration originate from the fields of psychology and neuroscience \cite{fleming2017hmeta, maniscalco2014signal}. They are typically applied in settings where a participant makes a primary judgment followed by a confidence rating—an assessment of the probability that the initial judgment was correct. Metacognitive sensitivity then refers to the individual's ability to distinguish correct from incorrect responses through confidence ratings \cite{galvin2003type}. On the other hand, metacognitive calibration refers to the extent to which these confidence ratings align with objective accuracy. For example, a confidence rating of 80\% with respect to a set of predictions is considered to be well calibrated if 80\% of those predictions are indeed correct. Importantly, one can certainly possess high metacognitive sensitivity while simultaneously exhibiting poor metacognitive calibration (or conversely). For instance, a participant might \textit{systematically} assign a confidence rating of 100\% to correct responses and a confidence rating of 99\% to incorrect ones. In such a case, metacognitive sensitivity would be perfect, as confidence ratings perfectly distinguish correct responses from incorrect ones. However, metacognitive calibration would be poor, as the participant is massively overconfident (systematically assigning a confidence level of 99\% to incorrect responses). Finally, it is important to stress that both dimensions are vital. Indeed, consider a scenario where an LLM communicates a confidence score regarding its own output. In this context, poor metacognitive calibration would cause the user to misinterpret the confidence score, whereas zero metacognitive sensitivity renders the confidence score entirely uninformative. 

Regardless, in what follows, our focus is only on the \textit{metacognitive sensitivity} of LLMs: the ability of LLMs to distinguish their correct responses from their incorrect ones. Existing studies have already explored the metacognitive sensitivity of LLMs \cite{li2025beyond, seo2026advice, cash2025quantifying}. However, in almost all these studies, the metrics used to quantify metacognitive sensitivity present a fundamental limitation (e.g., see Ref.~\cite{xiong2023can}). The reason for this is that in order to determine whether an agent has a ``good'', ``poor'', ``optimal'', ``sub-optimal'', ``better than'' or ``worse than'' metacognitive sensitivity, one must necessarily take into account \textit{cognitive sensitivity} (also known as type 1 sensitivity). Cognitive sensitivity—often quantified by $d'$—is the sensitivity associated with the primary task (i.e., the cognitive task targeted by confidence judgments). For the formal definition of $d'$, see the Methods section. The key point is that \textit{cognitive sensitivity determines the amount of information available for the subsequent confidence judgment task}. This statement can be derived from \textit{signal detection theory} (SDT) under minimal assumptions about how internal evidence is used for confidence judgments. On the one hand, this means that there is an upper bound on metacognitive sensitivity. Optimal metacognitive sensitivity is reached when confidence judgments are made using all the available information (having said this, in some specific cases, additional information can be accumulated between the primary task and the confidence judgment task, allowing metacognitive sensitivity to become higher than optimal). On the other hand, this means that metacognitive sensitivity is highly dependent on cognitive sensitivity. It is crucial to understand what this means. When two agents exhibit different levels of metacognitive sensitivity, this may be nothing more than a direct byproduct of their difference in cognitive sensitivity. In short, without taking cognitive sensitivity—$d'$—into account, one cannot know whether an observed difference in metacognitive sensitivity reflects a genuine difference in metacognitive abilities or merely a difference in cognitive sensitivity. 

Crucially, this problem is well known in the fields of psychology and neuroscience. In these areas, the gold standard metric for metacognitive sensitivity is called meta-$d'$ \cite{maniscalco2012signal, fleming2017hmeta, maniscalco2014signal, rahnev2025comprehensive}, and it has been extensively applied in numerous studies \cite{mazancieux2020metacognitive, rausch2015metacognitive, meunier2025does}. Meta-$d'$ has been developed specifically to overcome the problem discussed above in the context of binary tasks (see the Discussion section). Unlike the other measures of metacognitive sensitivity, meta-$d'$ shares the same units as $d'$. Remember that cognitive sensitivity—$d'$—determines the amount of information available for the subsequent confidence judgments task. By definition, meta-$d'$ is the $d'$ that an \textit{ideal observer} would need to generate the observed confidence ratings \cite{fleming2017hmeta}. As we can see, meta-$d'$ is a $d'$. It is precisely for this reason that meta-$d'$ and the true $d'$ (i.e., the one inferred directly from the data of the cognitive task) share the same unit, both being expressed in \textit{standard deviation} units. As such, meta-$d'$ can be compared directly with $d'$. To do so, one simply needs to compute the ratio meta-$d'$ divided by $d'$, known as $M_{ratio}$. Note that $M_{ratio}$ is a metric of \textit{metacognitive efficiency}, which can be defined as \textit{how much of the available information has been used to perform the confidence judgment task}. When $M_{ratio}$ equals 1, this means that all the available information has been used and that metacognitive sensitivity is optimal. On the other hand, when $M_{ratio}$ is strictly lower than 1, this means that only a fraction of the available information has been used and that metacognitive sensitivity is suboptimal. Furthermore, because meta-$d'$ rests on SDT, it remains unaffected by cognitive and metacognitive biases (at least in principle), as the very purpose of SDT is to dissociate \textit{sensitivity} from \textit{bias}. Having said this, the reader should be aware of some limit cases challenging the use of meta-$d'$. First, the long-held view that $M_{ratio}$ remains entirely decoupled from cognitive sensitivity and metacognitive bias has recently been challenged \cite{fleming2024metacognition, guggenmos2021measuring, xue2021examining, rausch2023measures, rahnev2025comprehensive}. Second, since this approach relies on SDT, one must assume that the underlying assumptions of SDT are valid, or at least sufficiently tenable to justify the use of meta-$d'$. This is why direct extensions of standard SDT have been proposed to relax specific assumptions \cite{miyoshi2026correcting, maniscalco2014signal}. Otherwise, it is worth noting that a ``model-free'' alternative to meta-$d'$ has recently been developed \cite{dayan2023metacognitive}.

In this paper, we argue for the use of the meta-$d'$ framework when assessing the metacognitive sensitivity of AI systems, whenever possible. While recent works have begun to explore this metric in the context of LLMs, its application remains fragmented \cite{cacioli2026llms}. Ref.~\cite{trinh2025metacognitive} applied it to model selection and Ref.~\cite{dai2026rescaling} investigated the impact of confidence scale design on metacognitive efficiency. Additionally, Ref.~\cite{wang2025decoupling} utilized a method closely related to meta-$d'$ but under an alternative terminology instead of adopting the established one, thereby further obscuring connections between the relevant fields.



By leveraging the properties of meta-$d'$, we are able to conduct a multi-faceted analysis across three axes of comparison. These comparison axes allow us to highlight the shortcomings of the popular metric type 2 AUROC (AUC2) \cite{clarke1959two} and demonstrate the practical relevance of meta-$d'$, especially when cognitive sensitivity varies. In the following list, we present these axes and discuss the insights they each provide:

\begin{enumerate}
\item \textbf{Compare an LLM to optimality.} This is essential to determine whether an LLM's metacognitive abilities have reached their ceiling or if there remains a significant margin for improvement.

\item \textbf{Compare different LLMs.} When it comes to integrating an LLM into a decision-making process, the question of selecting the most appropriate model always arises. As we have seen, an effective decision-making process must take uncertainty into account (particularly when risk is involved). Therefore, selecting the most appropriate model requires evaluating and comparing different LLMs not only on the cognitive task directly related to the decision-making process but also on their ability to quantify uncertainty. In the current paper, we investigated GPT-5 (reasoning efforts set to ``Medium'' by default \cite{openai_chatcompletions_create}), DeepSeek-V3.2-Exp (in non-thinking mode \cite{deepseek_api_doc}) and Mistral-Medium-2508. 

\item \textbf{Compare the same LLM in the context of different cognitive tasks.} This is important because if it turns out that metacognitive efficiency differs significantly depending on the targeted cognitive task, then it becomes difficult, if not impossible, to generalize conclusions regarding the LLM's metacognitive abilities established in the context of a specific cognitive task. Here, we consider three binary discrimination tasks, labeled A, B and C (see Methods for further details):

\begin{itemize}
\item \textbf{Sentiment analysis task (A).} The AI receives a sentence or piece of sentence, whose binary valence (either positive or negative) has been labeled by a human, and it must guess this valence.

\item \textbf{Oral vs written classification task (B).} The AI receives a sentence, that originates from either an oral or written source, and it must guess whether it is oral or written.

\item \textbf{Word depletion detection task (C).} The AI receives a sentence in which a ``the'' may have been deleted. The AI has to guess whether a ``the'' has been removed or if the sentence has been directly submitted unchanged.

\end{itemize}

\end{enumerate}

Note that the experimental setting at play here is always an LLM conducting a cognitive judgment (type 1 task) followed by a confidence rating—an assessment of the probability that the initial judgment was correct (type 2 task).


Finally, as mentioned above, in a second step, we explore the ability of LLMs to spontaneously adjust their decisions based on \textit{uncertainty} and \textit{risk configuration}. In principle, higher stakes are expected to yield more \textit{conservative} behavior. For instance, if the cost of failure with respect to a specific choice is catastrophic, then this path should only be considered when uncertainty is near zero. In the context of SDT \cite{green1966signal, hautus2021detection, wickens2001elementary}, such sensitivity to uncertainty and risk is captured by the shift of the decision criterion $c$. 


The question we ask is the following: depending on the risk configuration, does the LLM calibrate accordingly its type 1 criterion $c$? To answer this question, our experiments are straightforward: the AI performs a type 1 task (we use the same as those presented above) under different risk configurations. Calling S1 and S2 the two possible responses to the type 1 task, we refer to the risk configurations ``S1'' and ``S2'' the scenarios in which we use a prompt specifying high risk when answering S1 or S2, respectively, and ``None'' in case of no specification of risk. As previously, we consider GPT-5, DeepSeek-V3.2-Exp and Mistral-Medium-2508. See Methods for further details.

We refer to the experiments in which we estimate meta-$d'$ as the ``meta-$d'$ experiments'', while those focusing on the calibration of $c$ are termed ``$c$-calibration experiments'' (see Methods for further details).

\section{Results}\label{sec:results}

Given the large number of trials in our experiments, which can make even negligible effects statistically significant, we distinguish between statistical significance and \textit{practical} significance by defining \textit{Regions of Practical Equivalence} (ROPEs) \cite{makowski2019bayestestr}. For more information, see the Methods section. Hence, when a difference is said to be significant without further qualification, it means that it is \textit{both} statistically and practically significant. The meta-$d'$ experiments reveal that all three LLMs exhibited measurable metacognitive sensitivity, with $M_{ratio}$ values ranging from moderate to near-optimal, although some significant variation was observed across models and tasks. Crucially, we systematically compare the conclusions drawn from the meta-$d'$ approach with those drawn from the widely used metric of metacognitive sensitivity AUC2 \cite{clarke1959two}. In the $c$-calibration experiments, shifts in decision criterion $c$ indicate that all three LLMs were able to spontaneously adjust their decisions in a manner consistent with the specified risk.

\paragraph{Meta-$d'$ experiments}

\begin{figure}[h]
  \centering
  \includegraphics[width=\linewidth]{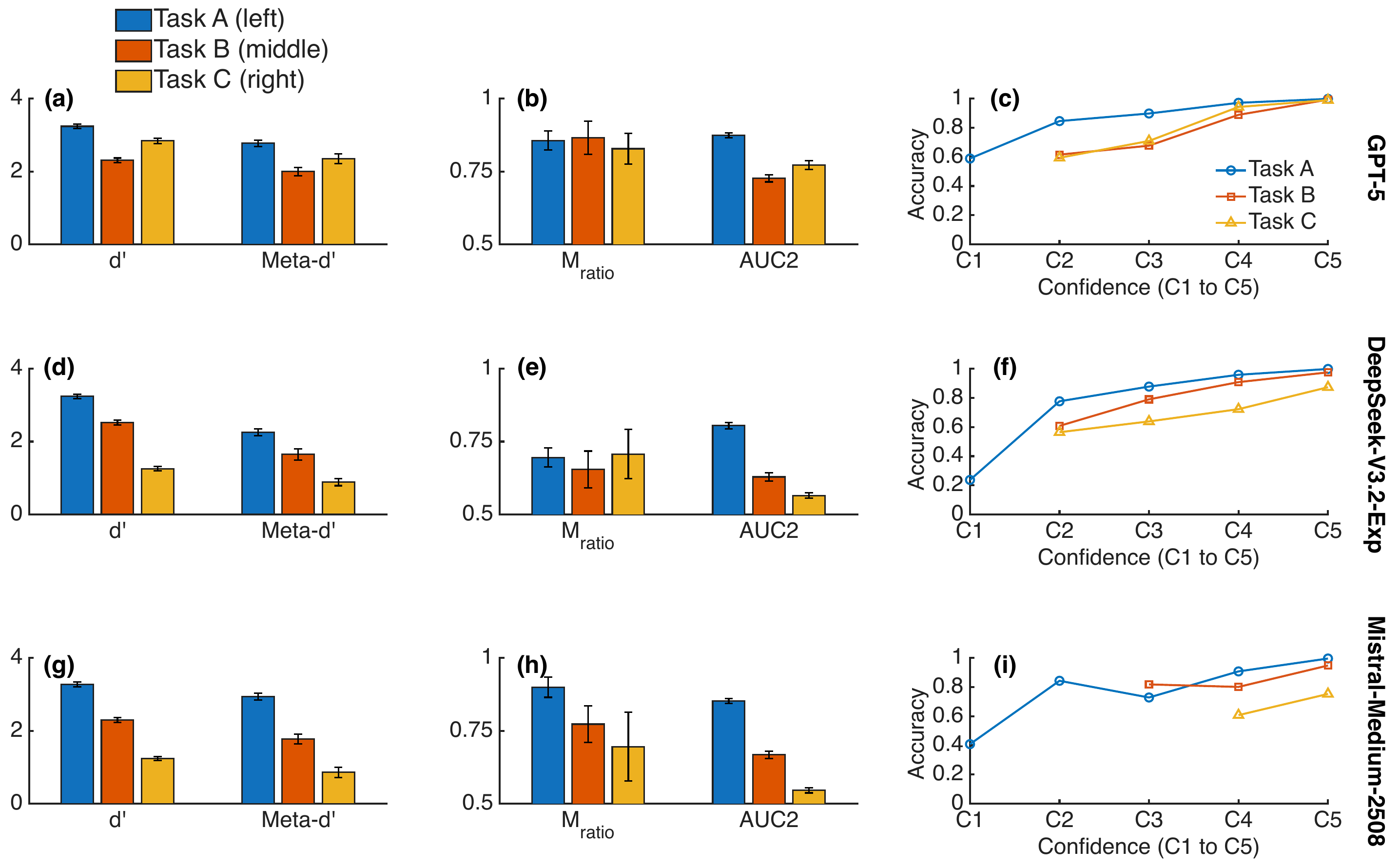}
  \caption{$d'$ and meta-$d'$ (left column), $M_{ratio}$ and AUC2 (middle column) and the accuracy, or $P(\text{correct} \mid \text{confidence} = C_i )$ versus confidence level (right column). The results are shown for GPT-5 (top row), DeepSeek-V3.2-Exp (middle row) and Mistral-Medium-2508 (bottom row) across task A (blue; left bars), task B (orange; middle bars), and task C (yellow; right bars). See the legends on top of panel (a) and in panel (c). Error bars in the left and middle columns denote 95\% CI of the posterior or estimated by the Delta method for AUC2 (see the Methods for details). Also, because the chance-level value of AUC2 is 0.5, and that all $M_{ratio}$ values are greater than 0.5, we set the range of the y-axis of the middle column to 0.5-1. When computing accuracy versus confidence (right column), GPT-5 reported confidence C1 in one trial of task B and one trial of task C; these observations were excluded, leaving no data points at those confidence levels. The number of trials was $2 \times 10^4$ for task A and $10^4$ otherwise (See the Methods for further details).}\label{fig:meta_d}
\end{figure}

The results of the meta-$d'$ experiments are shown in \Cref{fig:meta_d} and given in Table~S1. The left column of \Cref{fig:meta_d} shows $d'$, meta-$d'$, $M_{ratio}$, and AUC2, while the right column shows accuracy (proportion of correct responses) as a function of confidence ratings on a five-point scale. The first, second, and third rows correspond to GPT-5, DeepSeek-V3.2-Exp, and Mistral-Medium-2508, respectively.

We start by comparing the models' $d'$ across the tasks. In task A, no pairwise differences were observed (the 95\% Credible Interval [CI] over the posterior of their difference does not overlap zero; see the Methods section for details). In task B, DeepSeek-V3.2-Exp shows a significantly larger $d'$ than those of the other models, which do not differ significantly from one another. Lastly, in task C, GPT-5’s $d'$ is larger than those of the other models, which do not differ significantly from one another. Notably, in this latter case, the differences in $d'$ between GPT-5 and the other models is considerably more pronounced than in the other two tasks (GPT-5's type 1 sensitivity is $126\%$ and $130\%$ larger than those of DeepSeek-V3.2-Exp and Mistral-Medium-2508, respectively). 

Regarding differences in $d'$ across tasks for the same model, every model exhibits significant variations across tasks.

We now turn to confidence ratings. The right column of \Cref{fig:meta_d} shows that, in task A, unlike the other tasks, the models use the full confidence rating scale (see the distributions of confidence ratings in Figure~S1). In particular, in panel (i), we observe that Mistral-Medium-2508 essentially uses only two points of the confidence rating scale, which can occur in cases of poor metacognitive calibration, even in the context of optimal metacognitive sensitivity as discussed in the introduction. Also, we note that, for Mistral-Medium-2508 and DeepSeek-V3.2-Exp, the accuracy associated with the lowest confidence level (C1) in task A is lower than 0.5, corresponding to below-chance accuracy. Such below-chance accuracy can be explained by specific configurations of type 1 and type 2 criteria. Regardless, this confidence level only concerns 1.91\% and 0.76\% of trials, respectively.


In all cases, the models displayed substantial meta-$d'$ across tasks, but always suboptimal (the 95\% CIs of the posteriors over $\log M_{ratio}$ do not overlap 0 and fall entirely outside the defined ROPE; see the Methods for details). 

In task A, the $M_{ratio}$ of GPT-5 is significantly larger than that of DeepSeek-V3.2-Exp ($\Delta M_{ratio} = 0.16$, where $\Delta M_{ratio}$ denotes the difference in $M_{ratio}$). Mistral-Medium-2508's $M_{ratio}$ is larger than that of DeepSeek-V3.2-Exp ($\Delta M_{ratio} = 0.2$), and is not significantly different than that of GPT-5. In task B, GPT-5's $M_{ratio}$ is significantly greater than those of the other two models ($\Delta M_{ratio} = 0.21$ for DeepSeek-V3.2-Exp and $\Delta M_{ratio} = 0.09$ for Mistral-Medium-2508). However, the difference with Mistral-Medium-2508 is practically inconclusive. Mistral-Medium-2508's $M_{ratio}$ is also significantly larger than that of DeepSeek-V3.2-Exp, yet at a practically inconclusive level. In task C, GPT-5 shows an $M_{ratio}$ larger than that of the other models. However, these differences are practically inconclusive. The $M_{ratio}$ values of DeepSeek-V3.2-Exp and Mistral-Medium-2508 do not differ significantly.

For these comparisons, AUC2 closely follows $M_{ratio}$ (in tasks A and C, the differences in AUC2 between GPT-5 and Mistral-Medium-2508, and between Mistral-Medium-2508 and DeepSeek-V3.2-Exp are statistically significant but practically negligible), suggesting that the possible contamination by $d'$ does not drive the observed differences in AUC2 between models. This would be expected given that $d'$ remains relatively similar between models. That said, in task C, GPT-5's $d'$ is much larger than that of the other models. However, this gap does not distort the conclusion drawn from AUC2, since differences in metacognitive efficiency—tracked by $M_{ratio}$—already pointed in the same direction (i.e., GPT-5's $M_{ratio}$ is also higher than that of the other models).


Let us now turn to the differences across tasks for the same model which is the dimension where $d'$ vary the most. Across tasks, the $M_{ratio}$ of GPT-5 and DeepSeek-V3.2-Exp do not differ significantly. In contrast, their AUC2 differ significantly in a way that mirrors $d'$ (see \Cref{fig:meta_d}). Accordingly, AUC2 and $d'$ were highly correlated as anticipated (Spearman correlation: $r_s = 0.9 $, $p < 0.002$). For Mistral-Medium-2508, $M_{ratio}$ varies significantly between task A and task B, and between task A and task C, with $\Delta M_{ratio} = 0.13$ and $\Delta M_{ratio} = 0.20$, but not between task B and task C. All the corresponding pairwise values of AUC2 differ significantly.

Up to this point, our analyses have followed the standard scientific method by comparing conditions where only a single dimension varies at a time (either across models within the same task or across tasks for the same model). However, given that the primary goal of this paper is methodological, we illustrate the shortcomings of AUC2 by stressing a specific situation where we vary simultaneously the model and the task and where AUC2 and $M_{ratio}$ lead to opposite conclusions. Specifically, we compare DeepSeek-V3.2-Exp in Task A against Mistral-Medium-2508 in Task B. In this case, differences in both $M_{ratio}$ and AUC2 are statistically significant (a Bonferroni correction with $\alpha'=0.05/m$ and $m =36$ was applied to the AUC2 comparisons) \textit{but in opposite directions}: Mistral-Medium-2508 has a larger $M_{ratio}$ than DeepSeek-V3.2-Exp, but it has a smaller AUC2, consistent with its lower $d'$.

\paragraph{c-calibration experiments}

\begin{figure}[h]
  \centering
  \includegraphics[width=\linewidth]{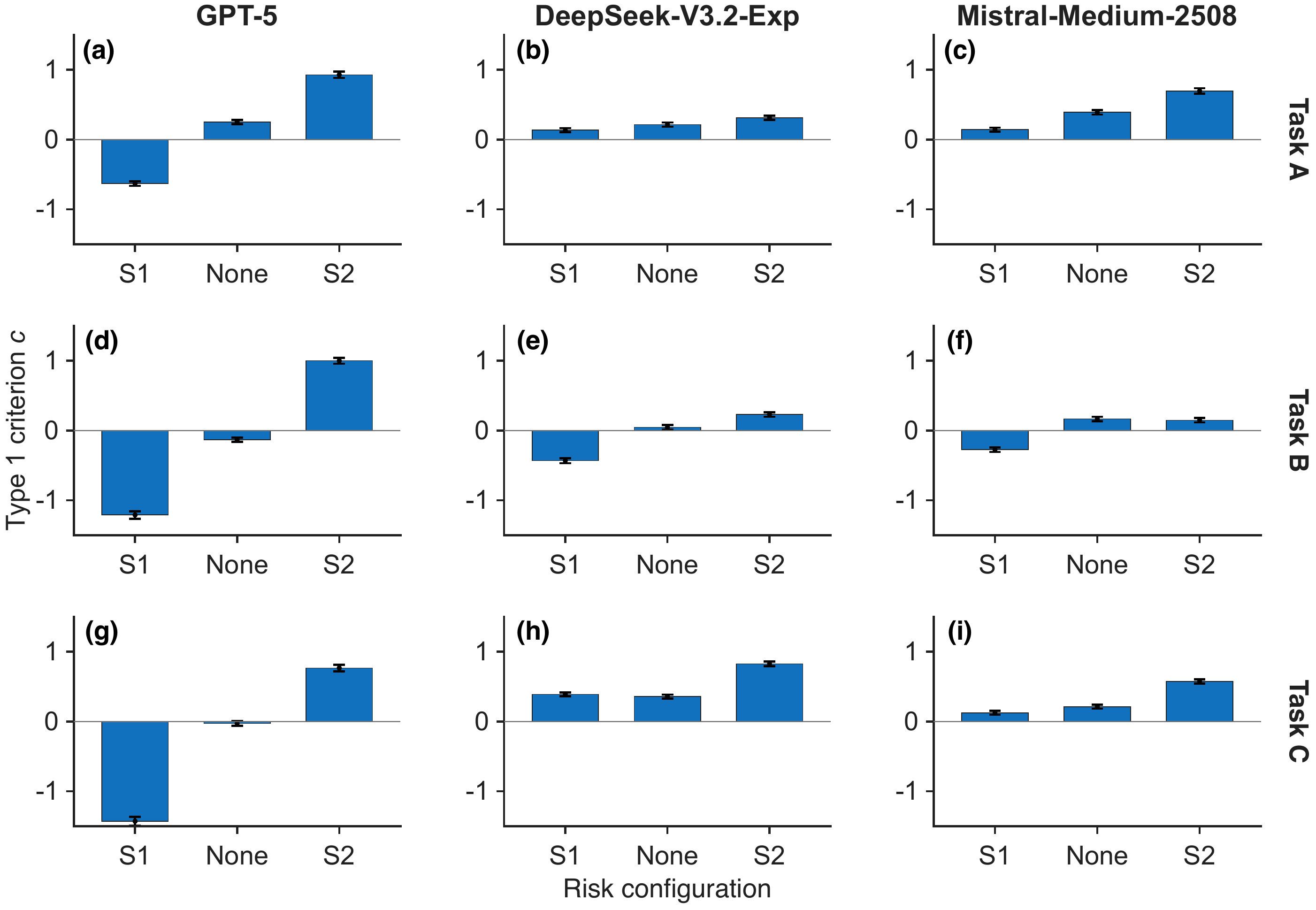}
  \caption{Type 1 criterion $c$ across three risk configurations (``S1'', ``None'' , ``S2'') within task A (top row), task B (middle row) and task C (bottom row), for GPT-5 (left column), DeepSeek-V3.2-Exp (middle column), and Mistral-Medium-2508 (right column). Error bars represent 95\% confidence intervals estimated via the Delta method. Although the Delta method can be sensitive to extreme proportions (that is, hit rates or false alarm rates close to 0 or 1), as may occur when responses are strongly biased toward S1 or S2, the large trial count ($N \gtrsim 10^4$) mitigates potential issues. See the Methods for details.}\label{fig:regulation_of_c}
\end{figure}

For all models and tasks, we vary the risk configuration (see Introduction and Methods for details). \Cref{fig:regulation_of_c} shows the estimated $c$ for each scenario, model and task (see also Table~S2). $c' = c / d'$ is given in Figure~S2.

GPT-5 is the only model to exhibit significantly different $c$ values across all risk configurations within all tasks. Importantly, these variations in $c$ values were logically coherent with the changes in risk configuration, as the model’s decision criterion consistently shifted to become more conservative toward the high-risk response (i.e., a shift to the left under the risk configuration ``S1'' and a shift to the right under the risk configuration ``S2''). Notably, under the risk configuration ``S1'', $c$ is always below $-0.6$, while $c$ under the risk configuration ``S2'' is always greater than $0.6$.

For Mistral-Medium-2508 within task B and for DeepSeek-V3.2-Exp within task C, the differences between the risk configurations ``None'' and ``S2'' and between the risk configurations ``S1'' and ``None'', respectively, are not significant. Otherwise, all other pairwise comparisons across risk configurations within each tasks are significant and consistent with the specified risk configuration. However, when compared to the defined ROPE (see Methods for details), the differences in task A between the risk configurations  ``S1'' and ``None'' and between the risk configurations ``S2'' and ``None''  for DeepSeek-V3.2-Exp are practically inconclusive. Similarly, in task C, the difference between the risk configurations  ``S1'' and ``None'' for Mistral-Medium-2508 is inconclusive.



It should be noted that several variations in $d'$ between risk configurations were found to be significant. Specifically, this is the case for GPT-5 within task B between the risk configurations ``None'' and ``S2'' (relative difference of $18\%$) and within task C between the risk configurations ``None'' and ``S1'' (relative difference of $41\%$) and between the risk configurations ``S2'' and ``S1'' (relative difference of $37\%$). Within task B, DeepSeek-V3.2-Exp also exhibits a significant change in $d'$ between ``S1'' and ``S2'' (relative difference of $13\%$). In contrast, all other differences were either non-significant (19 cases) or, while statistically significant, proved to be practically inconclusive (4 cases). 

When differences in $d'$ are significant, the normalized criterion $c' = c / d'$ is preferable to $c$, as it accounts for the magnitude of the bias \textit{relative to the sensitivity}. We therefore examine the differences in $c'$ for these specific cases in Figure~S2. Notably, the comparisons of $c'$ mirror those of $c$.


These results show that, in most cases across tasks and models, we were able to induce a conservative behavior toward either S1 or S2 by specifying a risk configuration in natural language through prompt design. In other words this demonstrates a general ability of LLMs to calibrate their type 1 criterion in a context-dependent manner.

\section{Discussion}\label{sec:discussion}


In this paper, we highlighted a key limitation of commonly used methods for assessing the metacognitive sensitivity of AI, and drew attention to the meta-$d'$ framework, a well-established approach from psychology and neuroscience specifically designed to address this shortcoming. Moreover, we also proposed to leverage SDT to measure the ability of AIs to spontaneously regulate their decisions based on uncertainty and risk. We showcased these psychophysical approaches by conducting two series of experiments on LLMs. In the first, our multi-faceted application of meta-$d'$ allowed us to compare LLMs to optimality, compare different LLMs on the same task, and compare the same LLM across different tasks, systematically contrasting the conclusions drawn from the meta-$d'$ approach with those drawn from the widely used metric AUC2 \cite{clarke1959two}. Notably, our results reveal that LLMs exhibit measurable metacognitive sensitivity. More precisely, all three models tested—GPT-5, DeepSeek-V3.2-Exp, and Mistral-Medium-2508—exhibited substantial values of meta-$d'$, indicating that their confidence judgments meaningfully tracked their type 1 performance, with variable $M_{ratio}$ levels, ranging from moderate ($\sim0.65$) to near-optimal ($\sim0.90$). Also, we found some variability in $M_{ratio}$ across tasks (for the same model) and across models (for the same task). Crucially, in some cases, AUC2 and $M_{ratio}$ lead to different—even opposite—conclusions, especially when large variations in type 1 sensitivity $d'$ contaminate the values of AUC2. This was found to be particularly the case when comparing the same model across tasks.

In sum, the results of the first series of experiments highlight the critical importance of the meta-$d'$ framework, particularly when cognitive sensitivity ($d'$) varies between conditions. That said, it should be noted that the core idea behind the meta-$d'$ approach—correcting metacognitive metrics for type 1 performance—can be extended to other measures of metacognitive sensitivity \cite{rahnev2025comprehensive, dayan2023metacognitive}. For example, Rahnev recently proposed the ratio of observed AUC2 to expected AUC2, where the latter is computed from estimated SDT parameters \cite{rahnev2025comprehensive}. Yet, after comparing several measures of metacognition on task performance and bias dependence, he concludes that $M_{ratio}$ remains a good default metric.

The second series of experiments, focused on shifts in $c$, highlighted the remarkable ability of LLMs to become more conservative in their decision-making in a way consistent with the specified risk. Hence, SDT constitutes a promising approach for studying the ability of AI systems to regulate their own decision.

We emphasize that our approach is strictly instrumental and outcome-oriented, making no claims about whether these systems possess genuine self-monitoring—i.e., metacognition—or merely mimic human behavior. While this distinction remains a fascinating question, it lies beyond the scope of this work. From a societal utility and risk-management perspective, what matters is merely the outcome (e.g., whether confidence ratings reliably predict type 1 performance), regardless of the underlying process that generates this outcome (on this topic, see Ref.~\cite{kumaran2026computational}). Note that there is a large literature on the mechanistic and computational differences between human brains and existing AI systems (e.g., see Ref.~\cite{connell2024can}).

Note that many aspects of our experiments can be tuned in a way that may affect the results. For example, instead of submitting the sentences one by one, we could submit them all at once, or in batches, to see whether this affects their responses as having access to all sentences could potentially impact their cognitive and metacognitive calibration. Similarly, for each trial, we could ask them to perform the type 1 task and the type 2 task in two successive prompts rather than in the same prompt. Also, it has been reported in Ref.~\cite{wang2025decoupling} that varying the design of confidence scale, as changing the number of confidence levels (e.g., $[0, 10]$ versus $[0, 20]$) affects metacognitive efficiency. One can also tune models' parameters such as the temperature or the level of ``reasoning budget''. Finally, remember that our article is not a benchmark paper.

Moreover, although our three tasks involved the discrimination of a single stimulus, psychophysical experiments are frequently carried out using two-alternative forced-choice (2AFC) discrimination tasks, in which the subject chooses between two alternative stimuli \cite{tyler2000signal} (we thank Kiyofumi Miyoshi and Hakwan Lau for valuable feedback on this point). The 2AFC paradigm is considered to render the equal variance assumption more tenable \cite{miyoshi2022assumptions}. In practice, the meta-$d'$ framework can also be applied beyond binary tasks (e.g., to open-ended questions) by
relabeling the responses to the cognitive task as ``correct'' or ``incorrect''. However, as it was specifically designed for binary
tasks, doing so may render some underlying assumptions untenable, which has led some authors to prefer not to use it in such settings \cite{cacioli2026llms, vuorre2022measures, molenberghs2016neural}. Regardless, whenever applicable, meta-$d'$ should be preferred. When it is not applicable, authors should exercise caution when drawing conclusions from comparisons across conditions in which type 1 performance and bias vary. As mentioned in the Introduction, there exist versions of SDT that relax specific assumptions, such as the unequal variance SDT model \cite{maniscalco2014signal}. Interestingly, unequal variance has been related to neuronal firing, where variance increases with the mean \cite{miyoshi2026correcting}.


Following standard practice in experimental psychology, we explicitly asked the AI—via the prompt—to report its confidence \cite{xiong2023can}; this is also a natural choice for \textit{conversational} AIs. An alternative would be to infer confidence from token probabilities \cite{steyvers2025large}. However, these two approaches need not be equivalent \cite{tian2023just, kumaran2026llms}. Note that verbalized confidence and token probabilities are two metrics among many different ones used to discriminate correct from incorrect responses in the field of LLM uncertainty quantification (UQ) \cite{fadeeva2023lm, shorinwa2025survey, ye2024benchmarking}.

We applied our approach in the context of simple binary discrimination tasks. However, it should be noted that the challenge of evaluating AI metacognition requires not only robust metrics but also the development of adequate tasks—specifically, tasks that actively engage the targeted metacognitive skills. For example, to evaluate the metacognitive competencies at stake in the ``AI decision-making'' scenario, we advocate for the use of experimental paradigms where the model must manage the exploration-exploitation trade-off (a dimension not addressed in our experiments). The classic paradigm for this is the multi-armed bandit problem \cite{lattimore2020bandit}. Another compelling example is the \textit{Eleusis Benchmark} recently proposed by David Louapre in Ref.~\cite{louapre2026}. In this task, the agent must find a hidden rule behind a sequence of cards. To do so, it samples evidence at each turn by playing a new card and receiving feedback, and then decides whether to commit to an answer. Crucially, the more delayed the successful guess and the more wrong answers, the less points. This creates a critical tension: responding too early leads to errors due to uncertainty, while responding too late costs points. While Louapre identified distinct ``player profiles''—ranging from reckless to cautious—our research adds a crucial layer to this observation: we have demonstrated that a single LLM is capable of strategic flexibility. Rather than being locked into one profile, these models can shift their behavior when the cost of error and context change. Finally, integrating such tasks with our psychophysical analysis would allow for a much finer characterization of AI metacognition. In ref.~\cite{kumaran2026causal}, the authors consider four-option multiple-choice questions plus an abstention option to investigate whether the AI chooses to abstain under low confidence. One possible extension of this setup would be to explicitly manipulate the level of risk specified in the prompt, in order to study whether the AI adjusts its abstention behavior in a context-sensitive manner.

We now turn to the well-known issue of hallucinations \cite{zhang2025siren}. Hallucinations are typically framed as factual errors communicated as \textit{undisputed} facts. Hence, hallucinations are not merely ``cognitive errors''; they also involve a ``metacognitive error,'' in that the model hallucinates a high confidence level in spite of uncertainty. The present article does not address this pathological regime.

While this study has focused on the ability of AI to estimate (and report) its own confidence, another crucial skill in some collaborative decision-making scenarios is the capacity to integrate confidence signals communicated by other agents—whether human or artificial. Future experiments will consist in providing an LLM with a recommendation accompanied by a confidence score whenever it is about to perform a type 1 task. The question we ask is: does the extent to which the AI takes into account the recommendation depends on the communicated confidence? Such experiments are crucial for evaluating whether AI systems can behave meaningfully when participating in collaborative decision-making processes. Note that a gap in the interpretation of confidence signals may arise between collaborating agents depending on how this information is formulated \cite{steyvers2025large}.

In conclusion, our work highlights the relevance of the meta-$d'$ and SDT approaches in the study of AI metacognition. Crucially, the growing integration of AI into decision-making workflows makes the characterization of AI metacognition an increasingly pressing need.

\section{Methods} \label{sec:methods}

\paragraph{Tasks and datasets}
In all experiments, the models are asked to perform a binary discrimination task (either the sentiment analysis task [A], the oral vs written classification task [B], or the word depletion detection task [C]; see below). In the meta-$d'$ experiments, they are also asked to rate their confidence on a scale from 1 to 5. The prompts are available on the GitHub page given below. Prompts for the $c$-calibration experiments may specify a risk configuration (see the Introduction for explanations). In all cases, we submitted one input (e.g., a sentence to discriminate) at a time, each submission effectively corresponding to a new conversation. \\

\noindent\textbf{Sentiment analysis task (A).} The AI receives a sentence or piece of sentence, whose binary valence (either positive or negative) has been labeled by a human, and it must guess this valence. We used the SST-2 dataset freely available at \url{https://huggingface.co/datasets/stanfordnlp/sst2}. \\

\noindent\textbf{Oral vs written classification task (B).} The AI receives a sentence, that originates from either an oral or written source, and it must guess whether it is oral or written. We used a labeled dataset freely available at \url{https://zenodo.org/records/7694423}.\\

\noindent\textbf{Word depletion detection task (C).} Using the same dataset from task B, we selected sentences comprising at least one ``the'', and deleted one with a probability of 0.5. The AI has to guess whether a ``the'' has been removed or if the sentence has been directly submitted unchanged.

\paragraph{Metrics}

We give a succinct introduction to type 1 sensitivity ($d'$), type 1 criterion ($c$), and metacognitive sensitivity (meta-$d'$) (see Ref.~\cite{maniscalco2014signal} for a thorough derivation). Suppose that a subject performs repeated trials of a discrimination task in which the only possible responses are S1 and S2. Given a stimulus, which is either S1 or S2, consider an internal response evaluated along a one-dimensional axis $x$. Let us define $f(x \mid S1)$ and $f(x \mid S2)$ to be normal distributions with equal unit variance. $d'$ is then defined as:

\begin{equation}\label{eq:d'}
    d' = z(\text{HR}) - z(\text{FAR})
\end{equation}
where $z$ is the inverse of the normal cumulative distribution function, HR is the hit rate (proportion of successful judgments when S2 was presented), and FAR is the false-alarm rate (proportion of incorrect judgments when S1 was presented). 
Notably, \Cref{eq:d'}  reduces to
\begin{equation}\label{eq:d_simplified}
    d' = \mu_{S2} - \mu_{S1}
\end{equation}
where $\mu_{S1}$ and $\mu_{S2}$ are the means of $f(x \mid S1)$ and $f(x \mid S2)$. $d'$ quantifies the extent to which the subject is able to discriminate the two types of stimuli—S1 and S2. Consider a decision criterion $c$ which is the value of the internal response $x$ below and above which the subject responds S1 and S2, respectively. It is possible to write $c$ as a function of HR and FAR:

\begin{equation}
    c = -0.5 \times \left(z(\text{HR}) + z(\text{FAR})\right).
\end{equation}
The smaller $c$, the stronger the internal response required for the subject to respond S1; that is, greater evidence is needed for a S1 response. We also define $c' = c / d'$.

After each trial, the subject is asked to rate its confidence on a given scale, from 1 to $H$. Meta-$d'$ is defined as the $d'$ that would be expected given confidence ratings only, if they were generated by an ideal observer \cite{fleming2017hmeta}. More generally, let us define $\theta = \left( \text{meta-}d', \text{meta-}c, \text{meta-}\textbf{\textit{\underline{c}}}_2  \right)$ where $\textbf{\textit{\underline{c}}}_2$ denotes the type 2 criteria \cite{maniscalco2014signal}. The estimation of meta-$d'$ is typically implemented using a maximum likelihood estimation:

\begin{align}
    \theta^* &= \underset{\theta}{\text{arg~max }} L(\theta \mid \text{confidence ratings})  \text{, subject to } \text{meta-}c' = c', \, \gamma\left(\text{meta-}\textbf{\textit{\underline{c}}}_{\text{ascending}} \right)
\end{align}
where $L$ is the log-likelihood and $\gamma\left(\text{meta-}\textbf{\textit{\underline{c}}}_{\text{ascending}} \right)$ is a Boolean function that returns true only if the type 1 and type 2 criteria are ordered appropriately. Metacognitive efficiency is defined as $M_{ratio} = \text{meta-}d' / d'$. It is possible to compute meta-$d'$ using a Bayesian model that outputs a full \textit{posterior}. In the current work, we used the implementation of this approach provided in Ref.~\cite{fleming2017hmeta}, available at \url{ https://github.com/smfleming/HMeta-d}. Yet, in the $c$-calibration experiments, we directly applied the formula of $c$ from type 1 response counts.

Finally, type 2 AUROC (AUC2) corresponds to the Area Under the type 2 Receiver Operating Characteristic curve. The type 2 receiver operating characteristic curve is obtained by varying the confidence criterion used to classify responses as high- or low-confidence and plotting, for each criterion, the proportion of correct responses classified as high-confidence against the proportion of incorrect responses classified as high-confidence.

\paragraph{Number of trials.}

For task A, the measured $d'$ were very large (above 3). Consequently, we submitted $2 \times 10^4$ sentences for each estimation. This trial count was informed by tests with simulated data (see Figure~S3). For all other tasks, the trial count was set to $10^4$. These large trial counts also support the use of the Delta method to quantify uncertainty in the estimates (see the next section).

\paragraph{Statistical analysis}

Following Ref.~\cite{fleming2017hmeta}, to assess statistical significance of differences between values of $d'$ and values of $M_{ratio}$ in the meta-$d'$ experiments, we checked whether the \textit{symmetric 95\% credible interval} (CI) of the posterior distribution of their logarithmic difference or log ratio overlap zero using Fleming's Bayesian model and specifically the function \texttt{fit_meta_d_mcmc.m} from Ref.~\cite{fleming2017hmeta} and freely available on the GitHub page associated with the paper, where we set \texttt{mcmc_params.estimate_dprime = 1}. To evaluate the optimality of the models' metacognition, we checked whether the CI of the posterior distribution of the log $M_{ratio}$ overlap zero. To assess the statistical significance of differences in AUC2, we checked whether $|Z| \geq 2.77$, where we applied a Bonferroni correction $\alpha'=0.05/m$ with $m = 9$ pairwise comparisons \cite{dunn1961multiple}, for a Z-test where we computed the variances of the differences using the Delta method \cite{cox2005delta}. Finally, for the statistical significance of differences in $c$ and $d'$ in the $c$-calibration experiments, we checked whether $|Z| \geq 3.11$, where we applied a Bonferroni correction $\alpha'=0.05/m$ with $m =27$ pairwise comparisons \cite{dunn1961multiple}, for a Z-test where we computed the variances of the differences using the Delta method \cite{cox2005delta}. 

Given the large number of trials in our experiments, which can make even negligible effects statistically significant, we defined Regions of Practical Equivalence (ROPEs) \cite{makowski2019bayestestr}. Hence, to assess the \textit{practical} significance of differences in $\log M_{\mathrm{ratio}}$ and AUC2, we use a ROPE of $[-0.05,\,0.05]$ (note that an AUC2 of 0.5 means that confidence ratings do not discriminate correct from incorrect responses better than chance while the corresponding chance-level value for $M_{ratio}$ is 0). For differences in $d'$ and $c$, we consider a ROPE of $[-0.1,\,0.1]$. In each case, we considered differences \textit{practically} significant if the 95\% CI fell entirely outside the ROPE, and on the contrary negligible if entirely inside, and inconclusive in the case of an overlap.

\section*{Code, software and data availability}

Our scripts, prompts, and results (data files) are freely available at

\noindent\url{https://github.com/sshrichard/metacognition-of-AI}

\section*{Author contributions}

The authors contributed equally to this work. P.S. and R.S. designed the project. R.S. prepared and conducted the experiments. R.S. and P.S. analyzed the results and wrote the manuscript.

\section*{Competing interests}

All authors declare no financial or non-financial competing interests. 

\section*{Funding declaration}

Not applicable

\bibliography{sn-bibliography}%

\newpage



\begin{center}
\section*{Supplementary material}
\end{center}
\setcounter{figure}{0}
\renewcommand{\thefigure}{S\arabic{figure}}
\setcounter{table}{0}
\renewcommand{\thetable}{S\arabic{table}}


\vspace{0.5cm}

\begin{figure}[h]
  \centering
  \includegraphics[width=\linewidth]{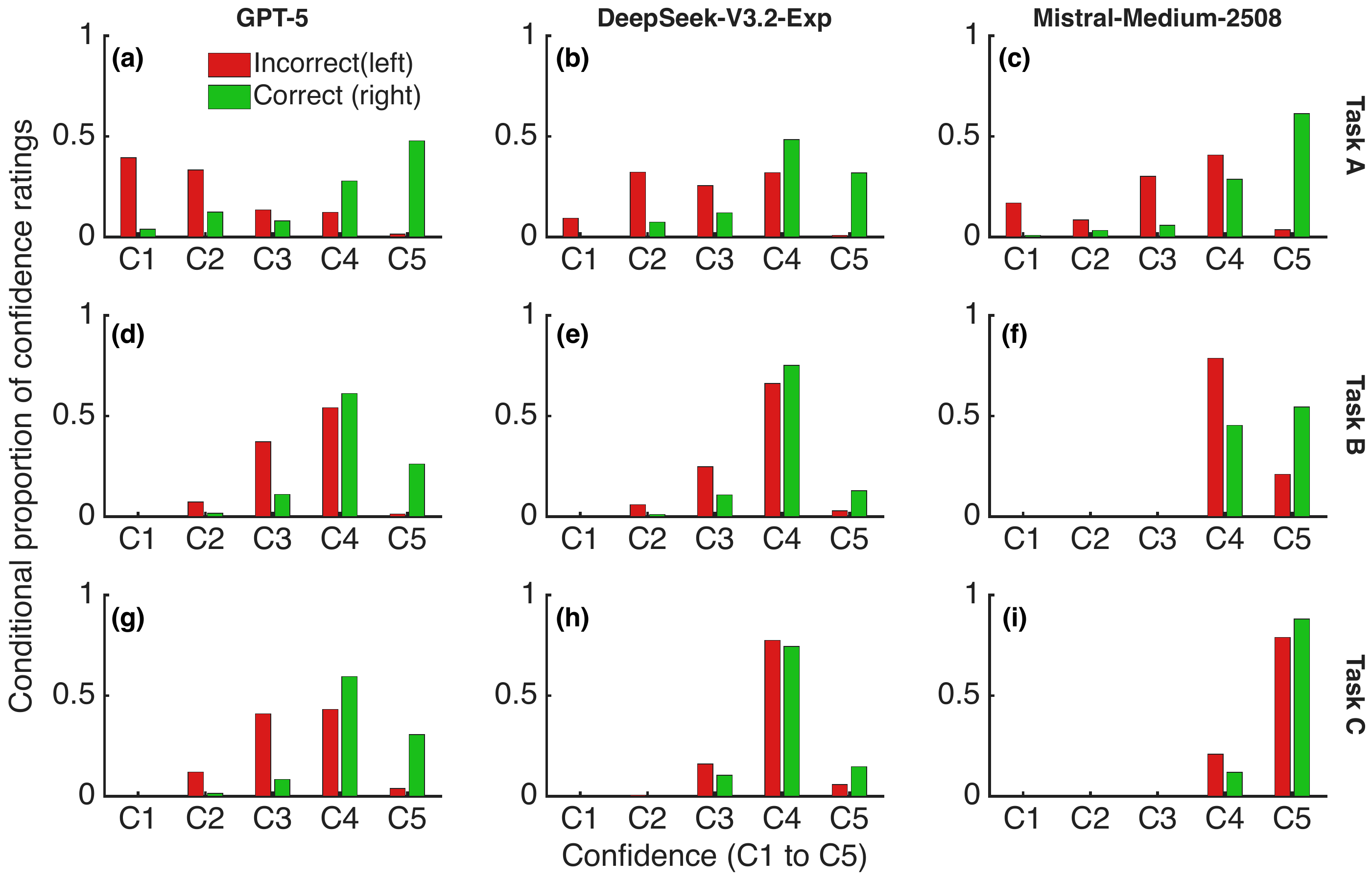}
  \caption{Conditional proportion of confidence rating given that the response at the type 1 task was correct (green; right bars) or incorrect (red; left bars), or $P(\text{confidence} = C_i \mid \text{correct or incorrect at type 1 task})$ for task A (first row), task B (middle row) and task C (bottom row). See \Cref{fig:meta_d} for further details. }\label{fig:dist_confidence}
\end{figure}

\begin{figure}[h]
  \centering
  \includegraphics[width=\linewidth]{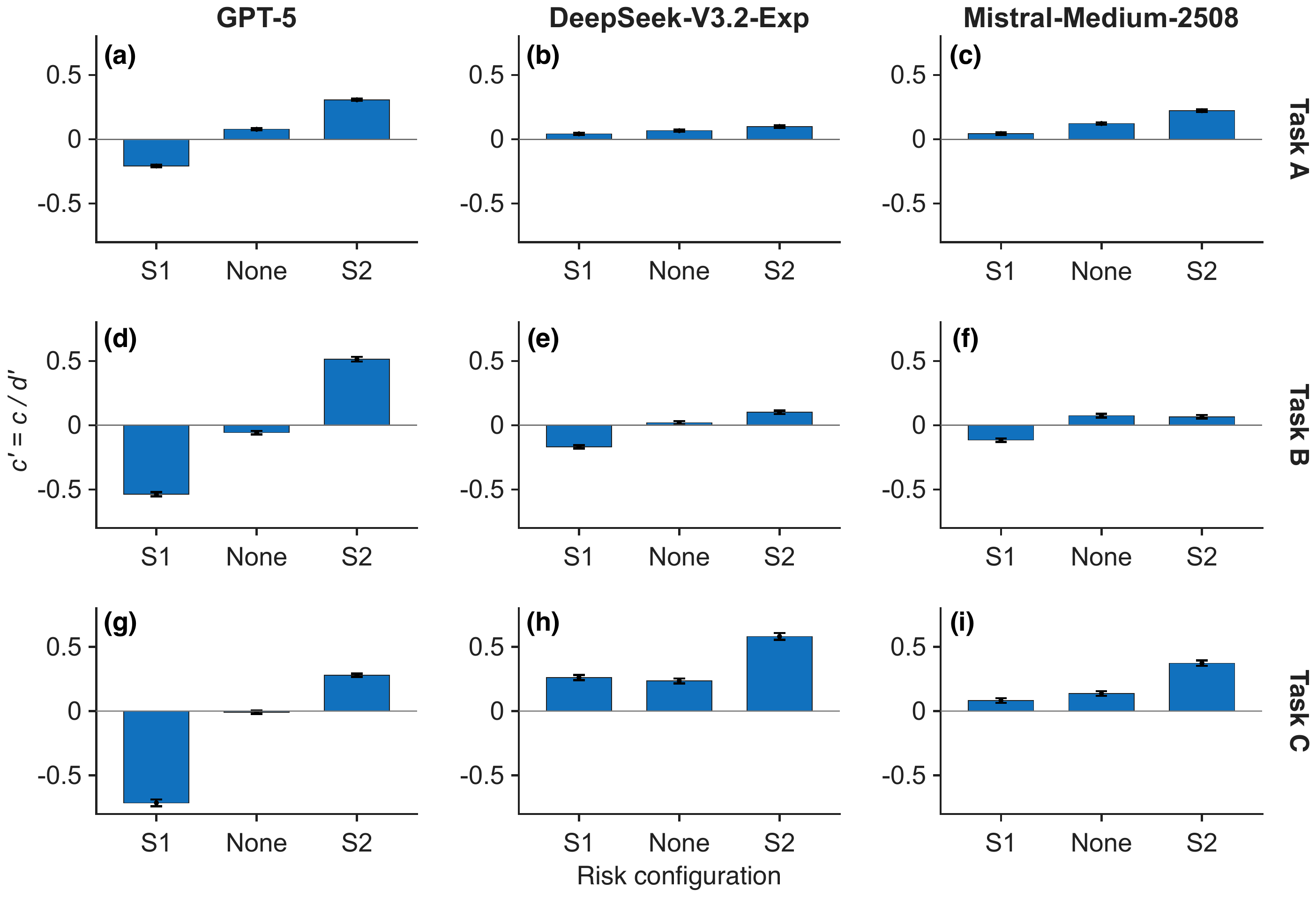}
  \caption{$c' = c / d'$ across three risk configurations (``S1'', ``None'' , ``S2'') within task A (top row), task B (middle row) and task C (bottom row), for GPT-5 (left column), DeepSeek-V3.2-Exp (middle column), and Mistral-Medium-2508 (right column). Error bars represent 95\% confidence intervals estimated via the Delta method. See the Methods for details.}\label{fig:c_prime}
\end{figure}

\begin{figure}[h!]
  \centering
  \includegraphics[width=0.9\linewidth]{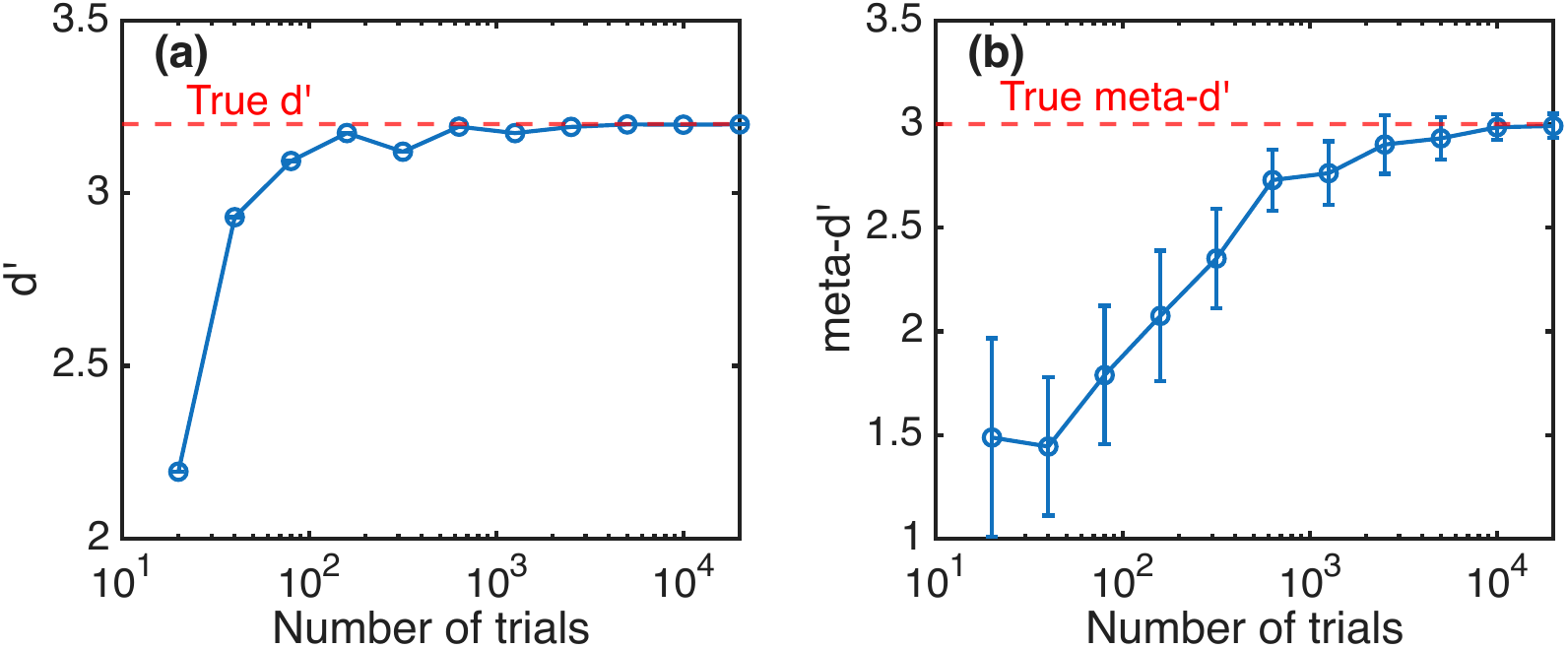}
  \caption{Average $d'$ (a) and meta-$d'$ (b) versus number of trials. Each data point is the estimated $d'$ and meta-$d'$, averaged over 20 repetitions, where each $d'$ and meta-$d'$ is computed with a given number of trials that is tuned in the x-axis. Error bars correspond to standard deviations. In panel (a), the error bars vanish because the simulation procedure deterministically set the type 1 response counts. The data has been generated using the function \texttt{exampleFit.m} from Ref.~\cite{fleming2017hmeta}, available at \url{https://github.com/smfleming/HMeta-d}, with the true $d'$ and meta-$d'$, indicated by a red horizontal dashed line in each panel, equal to 3.2 and 3 (i.e., very high type 1 sensitivity and high metacognitive efficiency), respectively. Type 1 criterion $c = 0$ and type 2 criteria are as follows: \texttt{c1  = [-2 -1.5 -1 -0.5]}, \texttt{c2  = [0.5 1 1.5 2]}.}\label{fig:synthetic_data}
\end{figure} 

\clearpage

\begin{table}[h]
\centering
\begin{tabular}{|c|c|c|c|c|c|c|c|}
\hline
\multicolumn{2}{|c|}{}
& \multicolumn{2}{c|}{GPT-5}
& \multicolumn{2}{c|}{DeepSeek-V3.2-Exp}
& \multicolumn{2}{c|}{Mistral-Medium-2508} \\
\hhline{~~------}
\multicolumn{2}{|c|}{}
& Estimation & $95\%$ CI
& Estimation & $95\%$ CI
& Estimation & $95\%$ CI \\
\hline
\multirow{3}{*}{Task A}
& $d'$        & 3.2396 & $[3.1776,\; 3.3031]$ & 3.2425     & $[3.1823,\;3.3040]$ & 3.2700  & $[3.2058,\;3.3359]$ \\
& meta-$d'$   & 2.7738 & $[2.6830,\;2.8627]$ & 2.2530 & $[2.1575,\;2.3499]$ & 2.9398 & $[2.8462,\;3.0334]$ \\
& $M_{ratio}$ & 0.8563 & $[0.8232,\;0.8889]$ & 0.6949 & $[0.6624,\;0.7283]$ & 0.8991 & $[0.8648,\;0.9344]$ \\
& AUC2 &0.8740& $[0.8659, \; 0.8821]$ & 0.8046 & $[0.7938, \; 0.8154]$ & 0.8517 & $[0.8433, \; 0.8600]$ \\
\hline
\multirow{3}{*}{Task B}
& $d'$        & 2.3074 & $[2.2442,\;2.3706]$ & 2.5217 & $[2.4553,\;2.5885]$ & 2.2968 & $[2.2343,\;2.3596]$ \\
& meta-$d'$   & 1.9972 & $[1.8772,\;2.1136]$ & 1.6510 & $[1.4939,\;1.8012]$ & 1.7743 & $[1.6385,\;1.9100]$ \\
& $M_{ratio}$ & 0.8658 & $[0.8094,\;0.9225]$ & 0.6548 & $[0.5911,\;0.7174]$ & 0.7727 & $[0.7104,\;0.8358]$ \\
& AUC2 &0.7263& $[0.7143, \; 0.7382]$ & 0.6288 & $[0.6145, \; 0.6432]$ & 0.6677 & $[0.6552, \; 0.6801]$ \\
\hline
\multirow{3}{*}{Task C}
& $d'$        & 2.8414 & $[2.7683,\;2.9148]$ & 1.2565 & $[1.1940,\;1.3192]$ & 1.2369 & $[1.1850,\;1.2887]$ \\
& meta-$d'$   & 2.3522 & $[2.2178,\;2.4886]$ & 0.8866 & $[0.7897,\;0.9824]$ & 0.8589 & $[0.7193,\;0.9990]$ \\
& $M_{ratio}$ & 0.8280 & $[0.7760,\;0.8805]$ & 0.7061 & $[0.6232,\;0.7925]$ & 0.6947 & $[0.5783,\;0.8134]$ \\
& AUC2 &0.7719& $[0.7565,\;0.7873]$ & 0.5651 & $[0.5560,\;0.5742]$ & 0.5455 & $[0.5369,\;0.5540]$ \\
\hline
\end{tabular}
\caption{Estimated $d'$, meta-$d'$, $M_{ratio}$ and AUC2} across models and tasks, and associated 95\% confidence intervals (CIs; see the Methods for details). Note that the $95\%$ CIs of differences mentioned in the main text should not be confused with the individual CIs.
\label{table:meta_d_values}
\end{table}

\begin{table}[h]
\centering
\begin{tabular}{|c|c|c|c|c|c|c|c|}
\hline
\multicolumn{2}{|c|}{}
& \multicolumn{2}{c|}{GPT-5}
& \multicolumn{2}{c|}{DeepSeek-V3.2-Exp}
& \multicolumn{2}{c|}{Mistral-Medium-2508} \\
\hhline{~~------}
\multicolumn{2}{|c|}{}
& Estimation & $95\%$ CI
& Estimation & $95\%$ CI
& Estimation & $95\%$ CI \\
\hline
\multirow{3}{*}{Task A}
& S1        & -0.6304 & ±0.0330 & 0.1356     & ±0.0291 & 0.1427  & ±0.0291 \\
& None   & 0.2532 & ±0.0308 & 0.2149 & ±0.0300 & 0.3927 & ±0.0324 \\
& S2 & 0.9289 & ±0.0452 & 0.3116 & ±0.0303 & 0.6975 & ±0.0388 \\
\hline
\multirow{3}{*}{Task B}
& S1        & -1.2098 & ±0.0537 & -0.4319 & ±0.0362 & -0.2748 & ±0.0328 \\
& None   & -0.1324 & ±0.0316 & 0.0465 & ±0.0324 & 0.1666 & ±0.0317 \\
& S2 & 1.0002 & ±0.0422 & 0.2301& ±0.0324 & 0.1487 & ±0.0318 \\
\hline
\multirow{3}{*}{Task C}
& S1        & -1.4297 & ±0.0644 & 0.3885 & ±0.0283 & 0.1243 & ±0.0273 \\
& None   & -0.0265 & ±0.0360 & 0.3584 & ±0.0283 & 0.2116 & ±0.0277 \\
& S2 & 0.7668 & ±0.0465 & 0.8260 & ±0.0323 & 0.5743 & ±0.0300 \\
\hline
\end{tabular}
\caption{Estimated $c$ across models, tasks, and risk configurations (``S1'', ``None'', ``S2'), and associated 95\% confidence intervals (CIs; see the Methods for details). }
\label{table:c_values}
\end{table}


\end{document}